%% file: main.tex
\documentclass[10pt,twocolumn,letterpaper]{article}

\usepackage[pagenumbers]{cvpr} 

\input{preamble}
\definecolor{cvprblue}{rgb}{0.21,0.49,0.74}
\usepackage[pagebackref,breaklinks,colorlinks,allcolors=cvprblue]{hyperref}
\usepackage{bm}
\usepackage{multirow}
\usepackage{makecell}
\usepackage{xcolor}
\usepackage{colortbl}
\usepackage{graphicx}

\def\paperID{} 
\def\confName{CVPR}
\def\confYear{2026}

\title{Revisiting 3D Reconstruction Kernels as Low-Pass Filters}

\author{Shengjun Zhang$^{1,*}$, Min Chen$^{1,*}$, Yibo Wei$^{2}$, Mingyu Dong$^{1}$, Yueqi Duan$^{1, \dag}$\\
$^{1}$Tsinghua University, $^{2}$Wuhan University \\
{\tt \small \{zhangsj23, cm22\}@mails.tsinghua.edu.cn, duanyueqi@tsinghua.edu.cn}}

\begin{document}
\maketitle
\newcommand\blfootnote[1]{%
\begingroup 
\renewcommand\thefootnote{}\footnote{#1}%
\addtocounter{footnote}{-1}%
\endgroup 
}
\blfootnote{\textsuperscript{*}Equal Contribution, \textsuperscript{\dag}Corresponding author.}

\input{sec/0_abstract}    
\input{sec/1_intro}
\input{sec/2_related_works}
\input{sec/3_signal_analysis}
\input{sec/4_ideal_splatting}
\input{sec/5_experiments}
\input{sec/6_conclusion}

{
    \small
    \bibliographystyle{ieeenat_fullname}
    \bibliography{main}
}

\newpage
\input{sec/X_suppl}

\end{document}

%% file: sec/0_abstract.tex
\begin{abstract}
3D reconstruction is to recover 3D signals from the sampled discrete 2D pixels, with the goal to converge continuous 3D spaces.
In this paper, we revisit 3D reconstruction from the perspective of signal processing, identifying the periodic spectral extension induced by discrete sampling as the fundamental challenge. 
Previous 3D reconstruction kernels, such as Gaussians, Exponential functions, and Student's t distributions, serve as the low pass filters to isolate the baseband spectrum.
However, their unideal low-pass property results in the overlap of high-frequency components with low-frequency components in the discrete-time signal’s spectrum.
To this end, we introduce Jinc kernel with an instantaneous drop to zero magnitude exactly at the cutoff frequency, which is corresponding to the ideal low pass filters.
As Jinc kernel suffers from low decay speed in the spatial domain, we further propose modulated kernels to strick an effective balance, and achieves superior rendering performance by reconciling spatial efficiency and frequency-domain fidelity.
Experimental results have demonstrated the effectiveness of our Jinc and modulated kernels.

\end{abstract} 

%% file: sec/1_intro.tex
\section{Introduction}
\label{sec:intro}

3D reconstruction~\cite{valkenburg1998accurate, triggs2000bundle, seitz2006comparison} is a fundamental task in computer vision, with broad applications ranging from autonomous driving to virtual reality. The goal of 3D reconstruction, to estimate the underlying 3D structures from the projected 2D images or videos, can be viewed as a specific instance of the classic signal processing problem to recover continuous signals from the sampled discrete ones. 

In this paper, we revisit 3D reconstruction through a signal processing lens, analyzing the problem in the frequency domain. 
The sampling of continuous 3D signals into discrete 2D pixels inevitably induces a periodic extension in the frequency spectrum, which poses a key challenge for 3D reconstruction. 
Guided by signal reconstruction theory, low-pass filters serve to suppress the aliased high-frequency components by isolating the baseband spectrum from its periodic replicas. Back in the spatial domain, this process is realized by a reconstruction kernel, defined as the inverse Fourier transform of the filter. 
The reconstruction is computed as a weighted superposition of the kernel, where a replica of the kernel is anchored at each sampling location.


By analogy with the signal theory, explicit 3D representations operate as reconstruction kernels, and their Fourier transforms as low-pass filters.
While traditional methods~\cite{MVS2007ICCV, SfM2016CVPR} leveraging point clouds and meshes lack visual fidelity for novel view synthesis~\cite{watanabe20253dgaborsplattingreconstruction}, 3D Gaussian Splatting (3DGS)~\cite{kerbl20233dgaussiansplattingrealtime} utilizes 3D Gaussian ellipsoids for explicit scene representations.
Benefiting from rasterization-based rendering, 3DGS avoids dense points querying in scene space, so that it can maintain high efficiency and quality.
Inspired by its remarkable success, following works~\cite{GES2024CVPR, zhu20253dstudentsplattingscooping, watanabe20253dgaborsplattingreconstruction} extend the reconstruction kernel from gaussian primitives to other representations.
For example, GES~\cite{GES2024CVPR} introduces Generalized Exponential Function to fit natural occurring signals, and 3D Student Splatting and Scoping (SSS)~\cite{zhu20253dstudentsplattingscooping} propose a mixture model consisting of Student's t distributions with both positive and negative densities.
Besides, 3DGabSplat~\cite{watanabe20253dgaborsplattingreconstruction} leverages a 3D Gabor-based primitive with multiple directional 3D frequency responses to enhance flexibility and efficiency.

\begin{figure*}
    \centering
    \includegraphics[width=\linewidth]{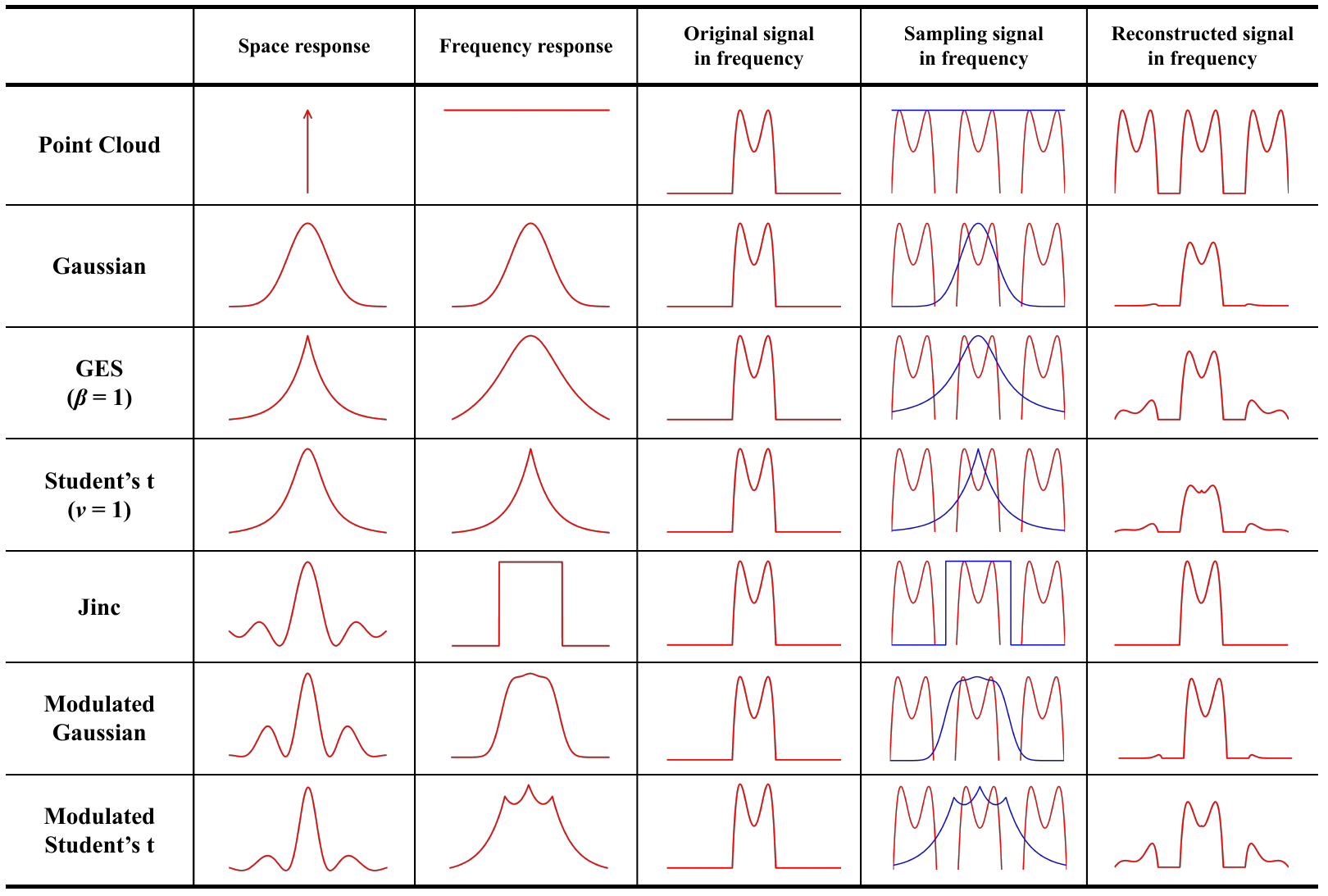}
    \caption{\textbf{Frequency analysis of 3D reconstruction kernels.} The table systematically summarizes key aspects for five kernel, including point clouds, Gaussians~\cite{kerbl20233dgaussiansplattingrealtime}, exponential function~\cite{GES2024CVPR}, student's t distribution~\cite{zhu20253dstudentsplattingscooping}, Jinc function and modulated kernels. This comparison aims to provide a qualitative reference for evaluating the spectral behavior of different kernels in 3D reconstruction, facilitating the design of optimal kernels for anti-aliasing, high-frequency information preservation, and low-frequency structural integrity.}
    \label{fig:teaser}
\end{figure*}

To validate these explicit 3D reconstruction kernels, we visualize the frequency response in Figure~\ref{fig:teaser}.
For point clouds, their frequency domain does not possess the filtering effect, which degrade the visual quality. 
Some other methods~\cite{kerbl20233dgaussiansplattingrealtime, GES2024CVPR, zhu20253dstudentsplattingscooping} have unideal property in the high-frequency regions, leading to the leakage of part of the high-frequency information.
To tackle this challenge, we introduce Jinc kernels, whose Fourier Transform is the ideal low-pass filters with the truncation in the frequency domain.
Based on this new representations, we propose a specifically rasterization-based rendering technique for Jinc kernels from Gaussian primitives to enable optimization and rendering.

Since Jinc kernels follow the $\propto r^{-1}$ decay speed in the spatial domain, an excessive number of kernels contribute to each pixel during rendering, leading to increasing memory and time consumption during rendering with higher resolutions.
In practice, we have to introduce premature truncation of the Jinc kernel in the spatial domain, but it breaks the continuous influence of the kernel across different tiles, resulting in rectangular artifacts.
Therefore, we further propose modulated kernels based on Gaussian functions and Student's t distributions, which inherit the fast spatial decay property of the original kernels, is closer to that of Jinc Splatting in frequency domain. 
In this manner, modulated kernels achieve superior rendering performance by reconciling spatial efficiency and frequency-domain fidelity.
Extensive experiments on real-world and synthetic scenes demonstrate that our methods outperform Gaussian primitives and other unideal reconstruction kernels.
Our main contributions can be summarized as follows:
\begin{enumerate}
    \item We revisit 3D reconstruction from the perspective of signal processing, and analyze the frequency property of explicit 3D kernels as the low-pass filters.
    \item We propose Jinc kernels, which is corresponding to ideal low-pass filters with an instantaneous drop to zero magnitude at the cutoff frequency.
    \item We introduce modulated kernels to benefit the spatial decay speed from based kernels, while achieving comparable performance to Jinc kernels in frequency domain. 
    \item Experiment results across both real-world and synthetic scenes, illustrate the efficacy of Jinc and modulated kernels in novel view synthesis.
\end{enumerate}



%% file: sec/2_related_works.tex
\section{Related Works}

\textbf{NeRF-based Novel View Synthesis.} Novel view synthesis (NVS) generates images from viewpoints different from those of the original captures~\cite{Lumigraph2023SGP, Lightfield2023SGP}. A key milestone is NeRF ~\cite{Mildenhall_2020_ECCV}, which parameterizes a continuous radiance field and renders images via volumetric integration along rays~\cite{RayTracing1990ToG, OpticalModels2002TVCG}. Subsequent works emphasize scalability and representation sharing across resolutions. Mip-NeRF~\cite{Barron_2021_ICCV, Barron_2022_CVPR} treats rays as conical frustums so a single network can model multi-scale structure. Some studies aim to solve the long-standing problem of novel view synthesis by improving the speed and efficiency of training and inference~\cite{EfficientNeRF2022CVPR, FastNeRF2021ICCV,InstantNGP2022TOG, KiloNeRF2021ICCV, NSVF2020NIPS,PlenOctree2021CVPR, Plenoxels2022CVPR}. 
Other research focuses on modeling complex geometry and view-dependent effects to reconstruct dynamic scenes~\cite{du2021neural,li2021neural,pumarola2021d,tretschk2021non,xian2021space}. 

\noindent\textbf{Primitive-based Differentiable Rendering.} Primitive-based approaches rasterize explicit geometric elements on the image plane and have gained traction for efficiency and controllability. 3D Gaussian Splatting (3DGS)~\cite{kerbl20233dgaussiansplattingrealtime} represents a scene as anisotropic 3D Gaussians and achieves high-definition real-time rendering through visibility culling and efficient rasterization. Building on this idea, later variants~\cite{zhu20253dstudentsplattingscooping, li20253dhgs3dhalfgaussiansplatting, GES2024CVPR, watanabe20253dgaborsplattingreconstruction} enrich the kernel family and the organization of primitives. For example, 3D Student Splatting and Scooping (SSS) ~\cite{zhu20253dstudentsplattingscooping} employs a heavy-tailed Student’s-$t$ kernel with positive/negative components for finer control, and GES~\cite{GES2024CVPR} introduces an expontial function to reduce the number of primitives.
Other methods~\cite{pixelSplat2023arXiv, MVSplat2024arXiv, SplatterImage2023arXiv, GPSGaussian2023arXiv, GGN2024NIPS} propose a feed-forward framework to learn powerful priors from large-scale datasets, so that 3D reconstruction and view synthesis can be achieved via a single feed-forward inference.

\noindent\textbf{Anti-Aliasing, Tomography and Optical Practice.}
Aliasing in view synthesis arises as a sampling-rate aliasing problem.
When scene frequencies exceed the Nyquist sampling rate, energy aliases unless prefiltering is applied prior to sampling. 
This principle classically motivates the use of ideal low-pass filters and their practical windowed variants~\cite{mitchell1988reconstruction, 1163711}.
It is echoed in contemporary footprint-aware anti-aliasing techniques for neural rendering, including Mip-NeRF~\cite{Barron_2021_ICCV}, Mip-NeRF360~\cite{Barron_2022_CVPR}, Zip-NeRF~\cite{Barron_2023_ICCV}, and Mip-Splatting~\cite{yu2023mipsplattingaliasfree3dgaussian}, as well as anti-aliased 3D Gaussian Splatting (3DGS)~\cite{liang2024analytic} and anti-aliased 2D Gaussian Splatting (2DGS)~\cite{younes2025anti} formulations, which analytically integrate over finite pixel footprints to mitigate aliasing.  
In tomography, the Radon transform and Fourier slice theorem underpin filtered back-projection techniques equipped with tailored frequency responses~\cite{radon20051}. Notably, recent computed tomography literature retains this filter-design perspective even when integrated with learning-based frameworks \cite{bhadra2022mining}.  
From an optics standpoint, a finite circular aperture acts as a built-in low-pass filter~\cite{goodman2005introduction}. Modern computational imaging and neural rendering works explicitly leverage this bandwidth-aperture relationship, as exemplified by DiffuserCam~\cite{antipa2017diffusercam}, LensNeRF~\cite{kim2024lensnerf}, and f-NeRF~\cite{hua2024fnerf}.

%% file: sec/3_signal_analysis.tex
\section{Signal Reconstruction Analysis}

Signal reconstruction typically commences with sampling the continuous original signal at a rate adhering to the Nyquist-Shannon criterion.
Subsequent to sampling, a Fourier transform is applied to convert these discrete spatial-domain samples into the frequency domain.
Next, a low-pass filter is employed to retain valid low-frequency components, while suppressing high-frequency noise and components exceeding the Nyquist limit, effectively avoiding high-frequency leakage and the mixing of high- and low-frequency information. 
Finally, an inverse Fourier transform is performed on the filtered frequency-domain data, converting it back to the spatial domain to yield a high-fidelity reconstructed signal that preserves the original signal’s key features while eliminating distortions caused by sampling.
We analyze the Fourier transform of classic kernel. For simplicity, we only consider 1D kernels.

\noindent\textbf{Point Clouds.}
Point clouds can be considered as Delta function, satisfying the sifting property $\int_{-\infty}^{+\infty} f(x)\delta(x-x_0) dx = f(x_0)$. Substituting into the Fourier transform definition and applying the sifting property with \( x_0 = 0 \), we obtain:
\begin{equation}
    {\mathcal{F}\{\delta(x)\} = 1},
\end{equation}
which indicates that points have no low-pass filtering ability in the frequency domain.

\noindent\textbf{Gaussian Function.}
For simplicity, the 1D Gaussian function has the form $g(x)=\exp\left(-\frac{x^2}{2\sigma^2}\right)$. The Fourier transform can be formulated as:
\begin{equation}
{\mathcal{F}\{{\exp\left(-\frac{x^2}{2\sigma^2}\right)}\} = \sqrt{2\pi}\sigma \exp\left(-\frac{f^2}{2(1/2\pi\sigma)^2}\right)},
\end{equation}
which is also a Gaussian function in the frequency domain.

\noindent\textbf{Exponential Function.}
The 1D exponential function is \( f(x) = A e^{-\alpha \left|x\right|} \). The transform simplifies to:
\begin{equation}
\mathcal{F}\{A e^{-\alpha \left|x\right|}\} = \frac{2\alpha A}{\alpha^2+f^2}.
\end{equation}

\noindent\textbf{Student's t-Distribution.}
The Student's t-distribution with degrees of freedom \( \nu > 0 \) and scale parameter \( \sigma > 0 \) is 
\begin{equation}
    f_\nu(x) = \frac{\Gamma\left( \frac{\nu+3}{2} \right)}{\Gamma\left( \frac{\nu}{2} \right) \pi^{\frac{3}{2}} \sigma^3} \left( 1 + \frac{x^2}{\nu \sigma^2} \right)^{-\frac{\nu+3}{2}},
\end{equation}
where \( \Gamma(\cdot) \) is the Gamma function. The Fourier transform is:
\begin{equation}
  \mathcal{F}\{f_\nu(x)\}=\frac{2^{1-\nu/2}}{\Gamma\left(\frac{\nu}{2}\right)}\left(\sqrt\nu\left|f\right|\right)^{\nu/2}K_{\nu/2}\left(\sqrt\nu\left|f\right|\right), \nonumber
\end{equation}
where $K(x)$ is modified Bessel function of the second kind.

%% file: sec/4_ideal_splatting.tex
\section{Methods}
In this section, we first introduce the Jinc kernels derived from the 3D ideal low-pass filter.
Based on this representation, we propose the Jinc splatting strategy for rasterization-based rendering.
Since Jinc kernels suffer from slow decay speed in the spatial domain, we further propose modulated kernels based on Gaussians and Student's t distribution to benefit from the spatial advantages of their base kernels, as well as comparable frequency response to Jinc kernels

\subsection{Jinc Kernel}
\label{subsec: Jinc Kernel}
Supposing that the ideal low-pass filter (ILPF) is a spherical indicator function that allows frequency components within a cutoff frequency $f_{c}$ to pass unattenuated while blocking a higher frequency, the formulation of 3D ideal low-pass filter is defined as:
\begin{equation}
    H(\bm{f}) = 
    \begin{cases} 
    1, & \|\bm{f}\| \leq f_c, \\
    0, & \|\bm{f}\| > f_c,
    \end{cases}
\end{equation}
where \(\|\bm{f}\| = \sqrt{f_x^2 + f_y^2 + f_z^2}\). This defines a spherical passband in the frequency domain.
The inverse Fourier transform becomes:
\begin{align}
    h(r) = & \int_0^{f_c} \int_0^{\pi} \int_0^{2\pi} e^{j 2\pi f r \cos\theta} f^2 \sin\theta \, d\phi \, d\theta \, df \nonumber\\
    = & \frac{1}{2\pi^2 r^3} \left[ \sin(2\pi f_c r) - 2\pi f_c r \cos(2\pi f_c r) \right] \nonumber\\
    = & \frac{2 f_c^2}{r} j_1(2\pi f_c r),
\end{align}
where $j_1(\alpha) = \frac{\sin \alpha - \alpha \cos \alpha}{\alpha^2}$ is the spherical Bessel function of the first kind.

We have discussed the isotropy ILPF above.
Now we introduce the definition of the anisotropy ILPF:
\begin{equation}
    h(\bm{x})=\dfrac{j_1(\sqrt{(\bm{x}-\bm{\mu})^{\top}\bm{\Sigma}^{-1}(\bm{x}-\bm{\mu})})}{\sqrt{(\bm{x}-\bm{\mu})^{\top}\bm{\Sigma}^{-1}(\bm{x}-\bm{\mu})}}, \label{eq:3D ILPF}
\end{equation}
where $\bm{\mu}$ is the central position of ILPF and $\Sigma=RSS^{\top}R^{\top}$ is the covariance matrix.
Here we ignore the cutoff-frequency, which can be absorbed into the scale matrix $S$.

\begin{figure*}[t]
  \centering
  \subfloat[Jinc Splatting]{\label{fig:sub_a}\includegraphics[width=0.245\linewidth]{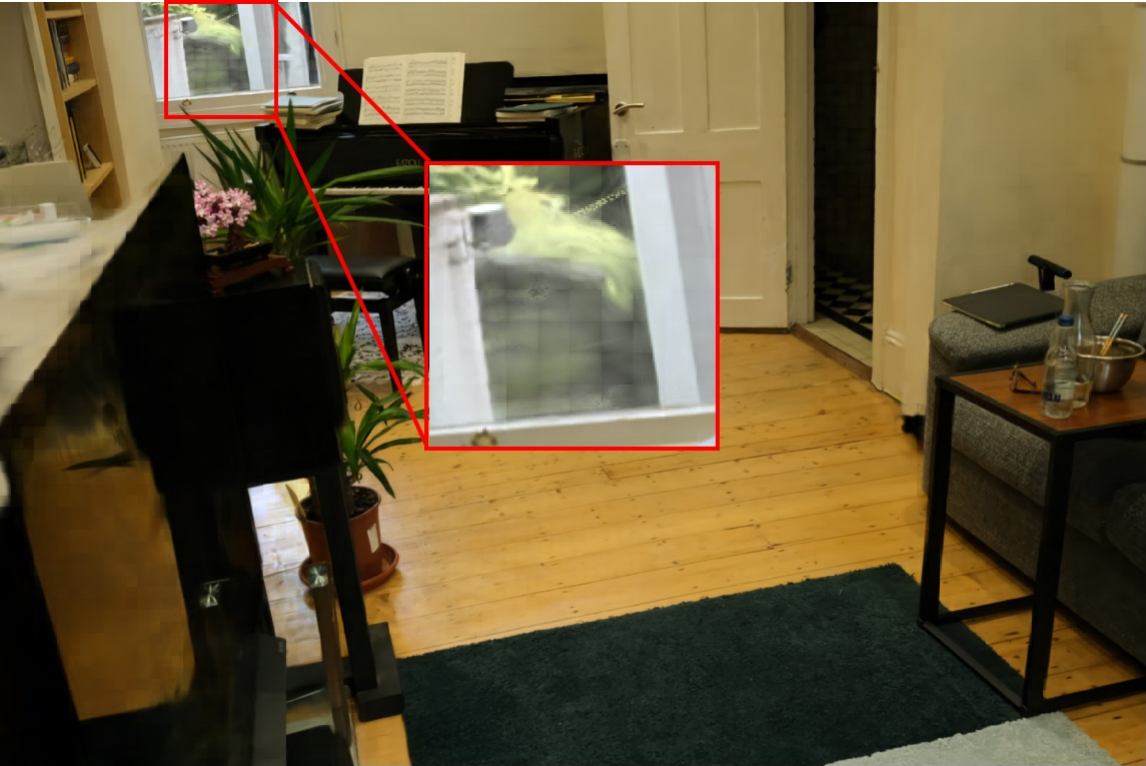}}
  \hfill
  \subfloat[Gaussian Splatting]{\label{fig:sub_b}\includegraphics[width=0.245\linewidth]{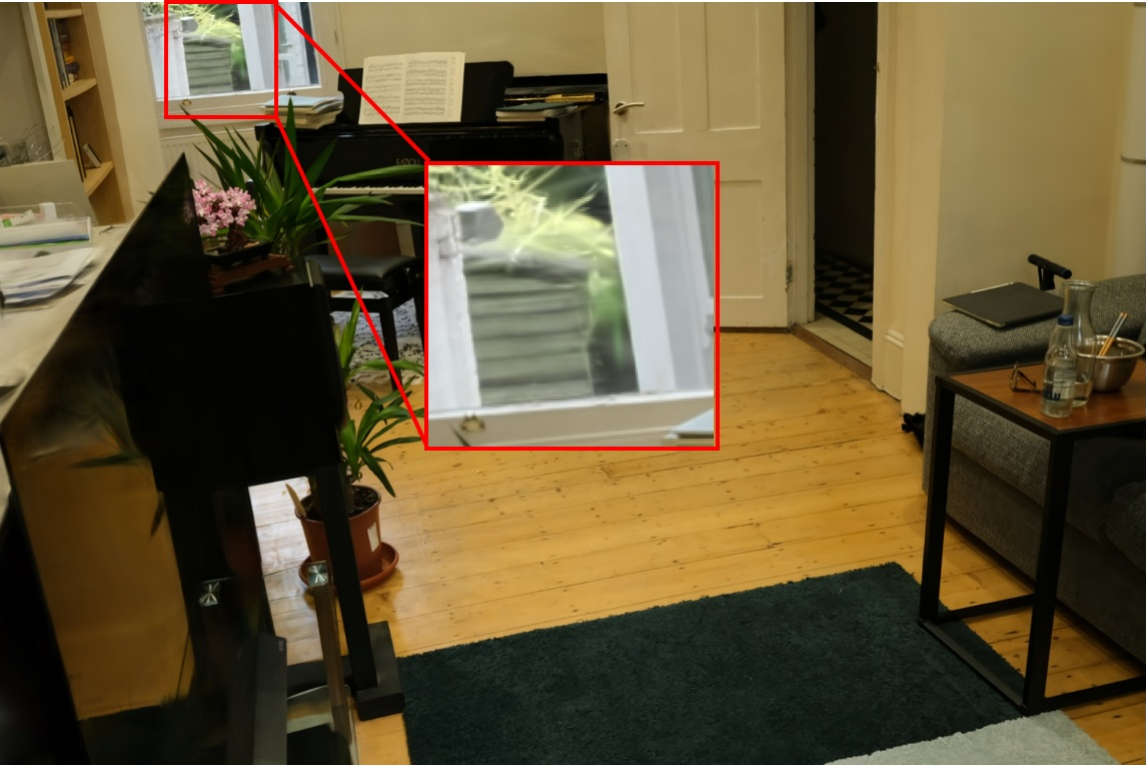}}
  \hfill
  \subfloat[Modulated Gaussian Splatting]{\label{fig:sub_c}\includegraphics[width=0.245\linewidth]{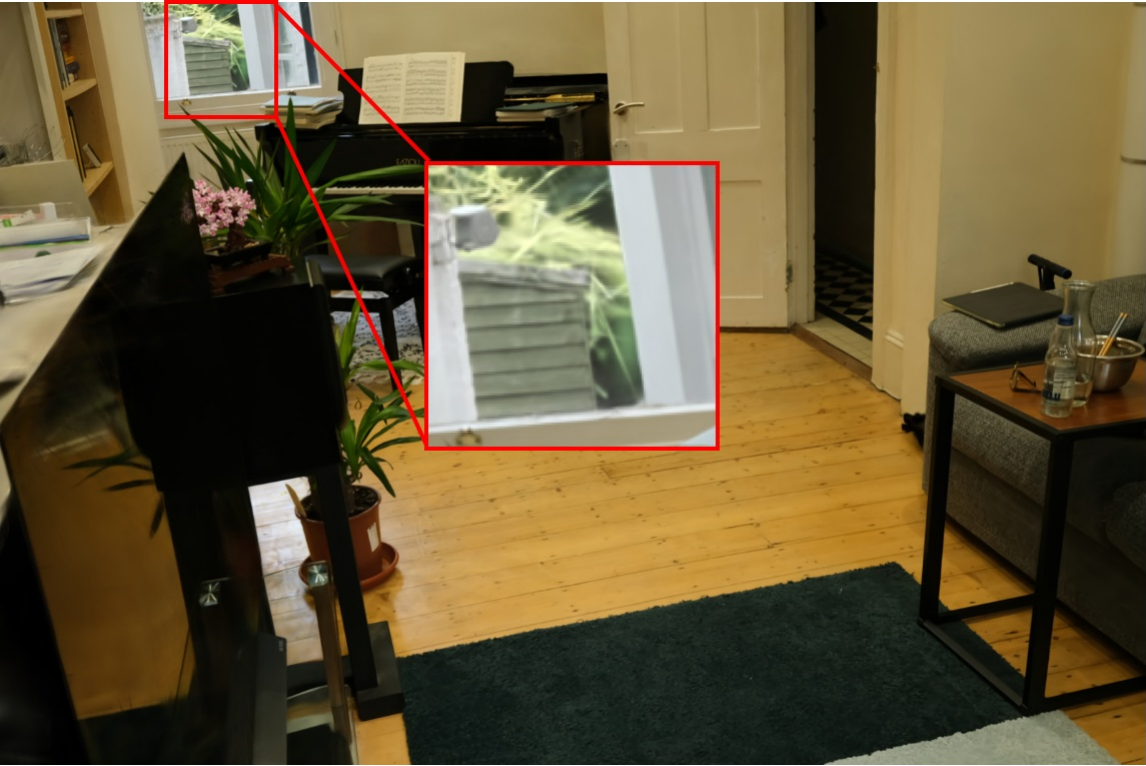}}
  \hfill
  \subfloat[Ground Truth]
  {\label{fig:sub_d}\includegraphics[width=0.245\linewidth]{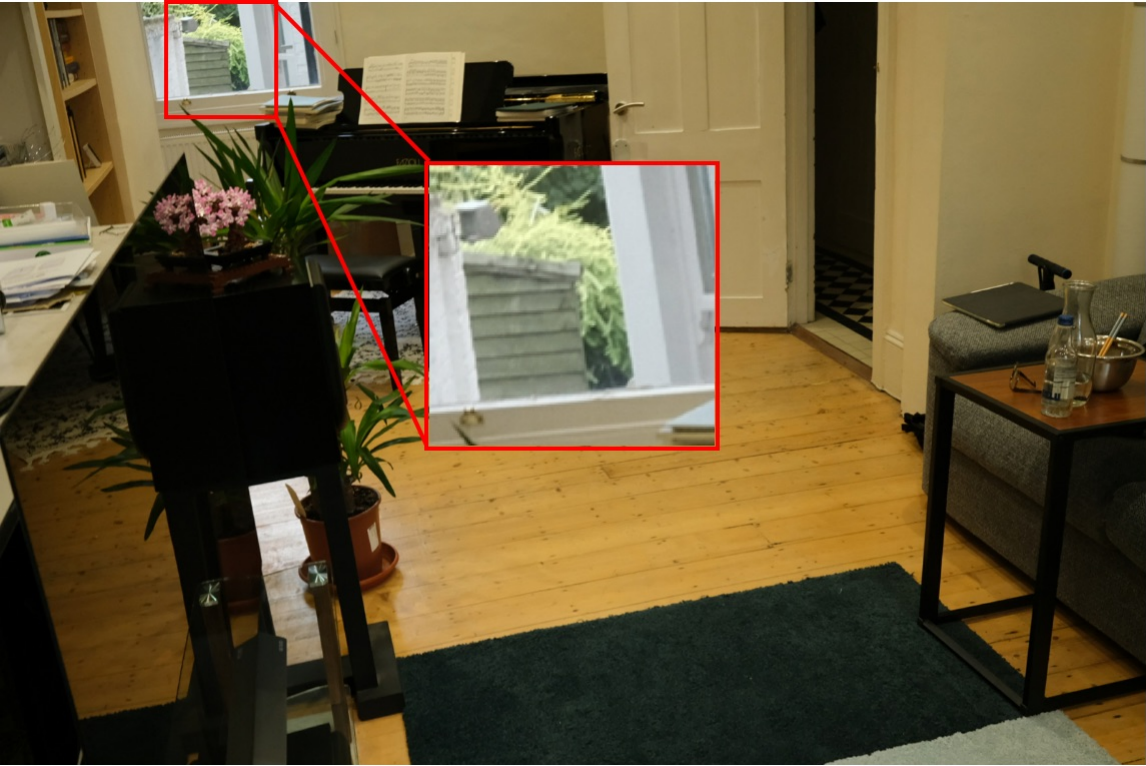}}
  
  \caption{\textbf{The influence of spatial decay speed.} Our Jinc-based method suffers from rectangular artifacts due to premature truncation, while Gaussians benefit from its rapid decay without such artifacts. Our Modulated Gaussian Splatting method strikes an effective balance, and achieves superior rendering performance by reconciling spatial efficiency and frequency-domain fidelity.}
  \label{fig:artifacts}
\end{figure*}

\subsection{Jinc Splatting}
\label{subsec: Jinc Splatting}
Rendering a pixel requires an affine transformation, then a projective transformation, followed by an integration along a ray.
Therefore, we first derive the integral of Jinc kernel along a line.
Then, we make the transformation for Jinc Splatting from the world coordinates to the pixel coordinates.
Finally, we discuss the active area when splatting a Jinc kernel in the image plane.

\subsubsection{Integral Along a Line}

This section derives the integral of the spatial response of a three-dimensional ideal low-pass filter along an arbitrary straight line \(\bm{x}(t) = \bm{a} + t \bm{b}\), where \(\bm{a} = (a_x, a_y, a_z)\), \(\bm{b} = (b_x, b_y, b_z)\), and \(t \in (-\infty, \infty)\).
We need to compute:
\begin{align}
    I & = \int_{-\infty}^{\infty} h(\bm{x}(t)) \, dt, \label{eq:integral}  \\
    & = \int_{-\infty}^{\infty} \dfrac{j_1(\sqrt{(\bm{x}(t)-\bm{\mu})^{\top}\bm{\Sigma}^{-1}(\bm{x}(t)-\bm{\mu})})}{\sqrt{(\bm{x}(t)-\bm{\mu})^{\top}\bm{\Sigma}^{-1}(\bm{x}(t)-\bm{\mu})}} \, dt, \nonumber\\
    & = \int_{-\infty}^{\infty} \dfrac{j_1\left(\sqrt{((\bm{a}-\bm{\mu})+t\bm{b})^{\top}\bm{\Sigma}^{-1}((\bm{a}-\bm{\mu})+t\bm{b})}\right)}{\sqrt{((\bm{a}-\bm{\mu})+t\bm{b})^{\top}\bm{\Sigma}^{-1}((\bm{a}-\bm{\mu})+t\bm{b})}} \, dt. \nonumber 
\end{align}
Since $\bm{\Sigma=RSS^{\top}R^{\top}}$, we denote $\bm{m=S^{-1}R^{-1}(\bm{a}-\bm{\mu})}$ and $\bm{n}=S^{-1}R^{-1}\bm{b}$. Thus, we have:
\begin{align}
   & ((\bm{a}-\bm{\mu})+t\bm{b})^{\top}\bm{\Sigma}^{-1}((\bm{a}-\bm{\mu})+t\bm{b}) \\
     = & \Vert \bm{m}\Vert^{2}+2t(\bm{m}\cdot\bm{n}) + t^2\Vert \bm{n}\Vert^{2}.
\end{align}
Substitute $s = t+\bm{m}\cdot\bm{n}/\Vert \bm{n}\Vert^{2}$, and the integration is transferred to:
\begin{align}
    I & = \int_{-\infty}^{\infty} \dfrac{j_1\left(\sqrt{((\bm{a}-\bm{\mu})+t\bm{b})^{\top}\bm{\Sigma}^{-1}((\bm{a}-\bm{\mu})+t\bm{b})}\right)}{\sqrt{((\bm{a}-\bm{\mu})+t\bm{b})^{\top}\bm{\Sigma}^{-1}((\bm{a}-\bm{\mu})+t\bm{b})}} \, dt, \nonumber\\
    & = \int_{-\infty}^{\infty} \dfrac{j_1\left(\sqrt{\Vert \bm{m}\Vert^{2}+2t(\bm{m}\cdot\bm{n}) + t^2\Vert \bm{n}\Vert^{2}}\right)}{\sqrt{\Vert \bm{m}\Vert^{2}+2t(\bm{m}\cdot\bm{n}) + t^2\Vert \bm{n}\Vert^{2}}} \, dt, \nonumber\\
    & = \dfrac{\pi J_1(\alpha)}{\Vert\bm{n}\Vert\alpha}, \label{eq:integral}
\end{align}
where $\alpha = \frac{\Vert\bm{m}\times\bm{n}\Vert}{\Vert\bm{n}\Vert}$.

\subsubsection{Coordinate Transformation}
The transform between the world coordinate system and the pixel coordinate system can be formulated as follows:
\begin{align}
\begin{bmatrix}
    u \\ v \\ 1
\end{bmatrix}
& = \begin{bmatrix}
    \bm{K} & \bm{0}
\end{bmatrix}\frac{1}{Z_c}
\begin{bmatrix}
    \bm{R}_c & \bm{t} \\ \bm{0}^\top & 1
\end{bmatrix}
\begin{bmatrix}
    X_w \\ Y_w \\ Z_w \\ 1
\end{bmatrix}.
\end{align}
Given the pixel $\bm{u}_h=[u,v,1]^{\top}$ in the pixel space, the integral direction in the world coordinate system is:
\[
\bm{d}_{w}=\bm{R}_c^\top\bm{d}_{c}=\bm{R_c}^\top\bm{K}^{-1}\bm{u}_h,
\]
where $\bm{d}_{c}$ is the ray direction in the camera coordinate.
The camera center in the world coordinate is given by $\bm{x}_{w}=-\bm{R}_c^\top\bm{t}$. Then, we get:
\begin{equation}
  \bm{x}(t)=\bm{a}+t\bm{b}=-\bm{R}_c^\top\bm{t}+t\bm{R}_c^\top\bm{K}^{-1}\bm{u}_h.  
\end{equation}
Thus, we get:
\begin{align}
    \bm{m}=-\bm{S}^{-1}\bm{R}^{-1}(\bm{R}_c^\top \bm{t}+\bm{\mu}), \quad
    \bm{n}=\bm{S}^{-1}\bm{R}^{-1}\bm{R}_c^\top \bm{K}^{-1}\bm{u}_h. \nonumber
\end{align}
Therefore, given the position of the pixel $(u,v)$, we can get the parameter $\alpha$ and the integral result.

\subsubsection{Radius of the Rendering Result}

Given the integral in (\ref{eq:integral}), we can set a threshold $q$. 
Once $\alpha$ exceeds $q$, we give the rendering result of ILPF a cutoff. 
We denote $\bm{n}=\bm{B}\bm{u}_h$ and get
\begin{align}
    q & = \frac{\Vert\bm{m}\times \bm{B}\bm{u}_h\Vert}{\Vert\bm{B}\bm{u}_h\Vert} \\
    & = \sqrt{\bm{m}^\top\bm{m}-\dfrac{\bm{u}_h^\top \bm{B}^\top \bm{m}\bm{m}^\top \bm{B}\bm{u}_h}{\bm{u}_h^\top \bm{B}^\top \bm{B}\bm{u}_h}}. \label{eq:radius}
\end{align}
We define $N=\bm{B}^\top \bm{m}\bm{m}^\top \bm{B}$ and $Q=\bm{B}^\top \bm{B}$ and simplify (\ref{eq:radius}) as:
\begin{equation}
    \bm{u}_h^\top(N-(\Vert\bm{m}\Vert^2-q^2)Q)\bm{u}_h=0.
\end{equation}
This is a homogeneous quadratic equation in 
$\bm{u}_hx$. Since $H=N-(\Vert\bm{m}\Vert^2-q^2)Q$ is a real symmetric matrix, the equation defines a conic section in the projective plane. This conic section is an ellipse if it satisfies $\text{det}(H)\neq0$ and $\text{det}(H_{2\times2})>0$, where $H_{2\times2}$ is the top-left $2\times2$ submatrix.
The ellipse can be written as:
\begin{align}
    & (\bm{x}-\bm{x_0})^\top{\bm{\Sigma}}_{2D}(\bm{x}-\bm{x_0})=1, \\
    & \bm{x_0}=-\bm{M}^{-1}\bm{p}, \quad {\bm{\Sigma}}_{2D}=-\dfrac{\text{det}(\bm{H}_{2\times2})}{\text{det}(\bm{H})}\bm{H}_{2\times2},
\end{align}
where $\bm{x}=[u,v]^\top$ is the coordinate in the image plane, $\bm{x_0}$ is the center of the ellipse, and $\bm{p}=[H_{1,3}, H_{2,3}]^\top$. The radius of rendering results is given by ${\Sigma}_{2D}$.

\subsubsection{Gradient Backward Propagation}
\label{subsubsec: Gradient}
According to (\ref{eq:integral}), the differential is given by:
\begin{equation}
    \frac{\partial I}{\partial\alpha} = \dfrac{1}{\Vert\bm{n}\Vert}\left[\dfrac{\pi J_0(\alpha)}{\alpha}-\dfrac{2\pi J_1(\alpha)}{\alpha^2}\right].
\end{equation}
Then, we compute the differential of $\alpha$ over $\bm{\mu}$ and $\Sigma$ for backward propagation to update these parameters.

By the chain rule for scalar-vector differentiation, we have
\begin{align}
    \frac{\partial \alpha}{\partial \bm{\mu}} & = \frac{1}{2\alpha}\frac{\partial (\alpha^2)}{\partial \bm{\mu}} \\
    & = \frac{1}{2\alpha}\left(\frac{\partial \Vert \bm{m} \Vert^2}{\partial \bm{\mu}}+\frac{\partial}{\partial \bm{\mu}} \left( \frac{(\bm{m}^\top \bm{n})^2}{\Vert \bm{n} \Vert^2} \right)\right) \\
    & = -\frac{1}{\alpha} (\bm{R S})^{-\top} \left( \bm{m} - \frac{\bm{m}^\top \bm{n}}{\Vert \bm{n} \Vert^2} \bm{n} \right)
\end{align}
Similarly, we have:
\begin{align}
    & \frac{\partial \alpha}{\partial \bm{\Sigma}} = \frac{\bm{\Sigma}^{-1}}{2\alpha (\bm{n}^\top \bm{\Sigma}^{-1} \bm{n})^2} \left( - (\bm{n}^\top \bm{\Sigma}^{-1} \bm{n})^2 \bm{m} \bm{m}^\top \right.\\
& \left. + 2 (\bm{m}^\top \bm{\Sigma}^{-1} \bm{n})(\bm{n}^\top \bm{\Sigma}^{-1} \bm{n}) \bm{n} \bm{m}^\top - (\bm{m}^\top \bm{\Sigma}^{-1} \bm{n})^2 \bm{n} \bm{n}^\top \right) \bm{\Sigma}^{-1}. \nonumber
\end{align}

\input{tab/decay_speed}

\subsection{Spatial Analysis}
\label{sec:spatial_analysis}
Although Jinc Splatting demonstrates superior filtering characteristics in the frequency domain, it exhibits slow decay in the spatial domain, resulting in an excessive number of kernels contributing to each pixel during rendering, which leads to increasing memory and time consumption during rendering. Under realistic GPU memory constraints, premature truncation of the Jinc kernel in the spatial domain is indispensable, which breaks the continuous influence of the kernel across different tiles, resulting in rectangular artifacts.
As shown in Figure~\ref{fig:artifacts}, when the Jinc kernel is used for high-resolution scene rendering, premature truncation in the spatial domain produces visible rectangular artifacts in the output images. 
Actually, the size of these rectangles matches the tile size employed in Gaussian Splatting. 
In contrast, the Gaussian kernel boasts fast spatial decay, whose influence fades rapidly within a limited spatial range, eliminating the need for excessive truncation, and thus inherently avoiding such artifact issues.

To this end, we evaluate the decay speed and energy concentration in both spatial domain and frequency domain comprehensively.
The decay speed classifies the kernel’s attenuation pattern, and the energy concentration is defined as the radial range containing 95\% of the kernel’s total energy.
The spatial energy concentration is defined as:
\begin{equation}
    \int_0^{r_{95\%}} r^2|h(r)|^2dr = 0.95\int_0^\infty r^2|h(r)|^2dr,
\end{equation}
where $r_{95\%}$ is the spatial range.
A small $r_{95\%}$ indicates that the kernel function's energy is highly concentrated in the central spatial region.
The energy concentration in frequency domain is similarly defined.
As show in Table~\ref{tab:decay speed}, Gaussian exhibits the fastest decay and the highest energy concentration in the spatial domain, while Jinc suffers from the slowest decay and the lowest concentration.
In the frequency domain: Jinc shows its superiority on decay speed, outperforming Gaussian and Student's t.

\begin{figure}
    \centering
    \includegraphics[width=\linewidth]{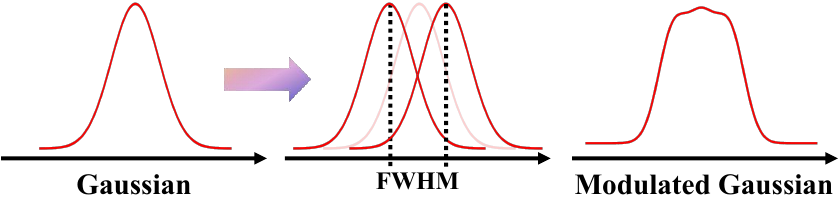}
    \caption{\textbf{Modulation of Gaussian kernels.} A proper frequency shift allows the Gaussian kernel to approximate an ideal filter.}
    \label{fig:frequency_modulation}
\end{figure}

\subsection{Modulated Kernels} 

To further demonstrate the effectiveness of our theoretical formulation while mitigating the slow spatial decay of the ideal reconstruction kernel, we embed our modulation strategy into two representative reconstruction kernels: the classical 3D Gaussian Splatting (3DGS)\cite{kerbl20233dgaussiansplattingrealtime}and the state-of-the-art 3D Student Splatting (SSS)\cite{zhu20253dstudentsplattingscooping}. 
This strategy is designed to retain the fast spatial decay and high energy concentration of the base kernels while reshaping their frequency response to approximate the ideal low-pass filter. 

Specifically, we perform frequency modulation on the reconstruction kernels by multiplying them by a cosine term, $\cos(\bm{f}^\top \bm{x})$, followed by an accumulation operation, which preserves the base kernel’s intrinsic spatial-domain efficiency while enhancing frequency-domain performance.
For simplicity, we take the 1D Gaussian kernel as an example. 
After applying modulation and accumulation, the resulting spatial-domain and frequency-domain expressions can be formulated as:
\begin{equation}
    h_g(x)=g(x)\left( \omega + (1-\omega)\text{cos}(f_0x) \right), \nonumber
\end{equation}
\begin{equation}
    \mathcal{F}(h_g(x))=\omega {G}(f)+\frac{1-\omega}{2}\left( {G}(f-f_0)+{G}(f+f_0) \right), \nonumber
\end{equation}
where $\omega \in [0,1]$ balances the original response and modulation of the kernel, and $f_0$ is the frequency shift. $g(x)$ and $\bm{G}(f)$ are the spatial and frequency-domain expressions of the 1D Gaussian kernel, respectively. 
For the Student’s t kernel, the modulation follows the same logic:
\begin{equation}
    h_t(r) = t(r)\left( \omega + (1-\omega)\text{cos}(f_0r) \right), \nonumber
\end{equation}
\begin{equation}
    \mathcal{F}(h_t(r)) = \omega T(f) + \frac{1-\omega}{2}\left( T(f-f_0)+T(f+f_0) \right), \nonumber
\end{equation}
where $t(r)$ is the Student's t kernel and $T(f)$ is its Fourier transform.

As illustrated in Fig.~\ref{fig:frequency_modulation}, we set the frequency shift parameter $f_0$ to half of the Full-width at half maximum (FWHM) of the base kernel. 
For a Gaussian function \( g(x) = A \exp\left(-\frac{(x-\mu)^2}{2\sigma^2}\right) \). The FWHM is given by: 
\begin{equation}
    \text{FWHM}_g = 2\sigma\sqrt{2\ln2} \approx 2.355\sigma.
\end{equation}
Thus, $f_0$ is set to $1.178\sigma$.
Similarly, the FWHM of Student's t kernels ($\nu=1$) is:
\begin{equation}
    \text{FWHM}_t =  2\sigma\ln 2 \approx 1.386\sigma
\end{equation}
As validated by Table~\ref{tab:decay speed}, modulated Gaussian kernels retains the base Gaussian’s exponential decay ($\propto e^{-r^2}$) with a 95\% energy concentration of ~3.4 which is marginally larger than 3.2, due to trivial oscillations suppressed by fast decay, while exhibiting narrower 95\% energy concentration in the frequency domain.
The phenomenon is also observed in the modulated Student's t kernel.
To this end, the modulation strategy enables Gaussian and Student's t kernels to inherit the spatial advantages of their base kernels, while achieving frequency-domain performance comparable to Jinc, as a trade-off between efficiency and anti-aliasing.


%% file: tab/decay_speed.tex
\begin{table}[t]
\centering
\caption{\textbf{Spatial and frequency traits of reconstruction kernels.} We report the decay speed and the 95\% energy range, whose units is normalized spatial length or wavenumber. For simplicity, we set $\nu=1$ for Student's t distribution.
}
\vspace{-0.2cm}
\label{tab:decay speed}
\resizebox{\linewidth}{!}{\begin{tabular}{l|cccc}
\toprule
\multirow{2}{*}{Method} & \multicolumn{2}{c}{Spatial Domain} & \multicolumn{2}{c}{Frequency Domain} \\ \cmidrule{2-5}
                        & Speed & Energy & Speed  & Energy  \\ \midrule
Gaussian~\cite{kerbl20233dgaussiansplattingrealtime}    &  $\bm{\propto e^{-r^2}}$    &    \textbf{2.77}  &   $\propto e^{-f^2}$    &    2.77                  \\
Student's t~\cite{zhu20253dstudentsplattingscooping} &  $\propto r^{-2}$     &    3.68  &   $\propto e^{-f}$  &    2.99                   \\\midrule
Jinc        &  $\propto r^{-1}$              &    5.59  &  \textbf{Cutoff}        &    \textbf{1.90}       
\\ 
Modulated Gaussian & $\bm{\propto e^{-r^2}}$ & $\sim$2.77 & $\propto e^{-f^2}$ & $<$2.77 \\
Modulated Student's t & $\propto r^{-2}$ & $\sim$3.68 &  $\propto e^{-f}$ & $<$2.99 \\
\bottomrule
\end{tabular}
}
\vspace{-0.3cm}
\end{table}

%% file: sec/5_experiments.tex
\section{Experiments}

\subsection{Low Resolution View Synthesis}

\input{tab/low_resolution}

\input{tab/full_resolution}

\textbf{Datasets and Matrices.} As demonstrated in Section~\ref{sec:spatial_analysis}, high-resolution rendering with our Jinc Splatting kernel incurs substantial memory usage and computational time due to its slow spatial decay. This makes direct evaluation under standard high-resolution settings impractical.
To enable a fair assessment of the method’s behavior, we construct low-resolution variants of the NeRF-Synthetic dataset~\cite{Mildenhall_2020_ECCV} at $64\times 64$ and $128\times128$ spatial resolutions. These datasets are generated by downsampling the original high-resolution images using lanczos interpolation, which preserves global structures and view-dependent appearance while reducing memory and computational demands.
We evaluate the full model of our method on eight synthetic scenes from the low-resolution NeRF Synthetic Dataset. Also, following the previous evaluation metrics, we use Peak Signal-to-Noise Ratio (PSNR), Structural Similarity Index Metric (SSIM)~\cite{SSIM2004TIP}, and Learned Perceptual Image Patch Similarity (LPIPS)~\cite{LPIPS2018CVPR}.

\noindent\textbf{Baselines.} We selected different reconstruction kernels as baselines for comparison, including the original 3D Gaussian Splatting (3DGS)~\cite{kerbl20233dgaussiansplattingrealtime}, and the state-of-the-art 3D Student Splatting (SSS)~\cite{zhu20253dstudentsplattingscooping}, which models the kernel using Student’s t-distribution.

\noindent\textbf{Implementation Details.} Our method is implemented on top of the 3D Student Splatting (SSS)~\cite{zhu20253dstudentsplattingscooping} codebase. We follow SSS in using the SGHMC optimizer and its adaptive density control strategy. Unlike Gaussian splatting methods that rely on the empirical $3\sigma$ rule to define kernel range, our Jinc kernel adopts the parameter $\alpha$ in Eq.~\ref{eq:integral} to specify its effective spatial range, which we fix to $\alpha = 30$ in all experiments. All experiments are conducted on a single NVIDIA A800 GPU. For the $64\times64$ setting, we train for 10,000 iterations due to the reduced scene complexity, while for the $128\times128$ setting, we extend training to 15,000 iterations to ensure adequate convergence. These configurations provide consistent and fair comparisons across different kernel designs and resolutions.

\noindent\textbf{Main Results.}
Table~\ref{tab:low resolution result} presents the performance of different reconstruction kernels in low-resolution novel view synthesis. 
At $64\times64$ resolution, the Jinc kernel achieves comprehensive superiority, where it reaches a PSNR of 29.87 dB, representing a 0.70 dB improvement over SSS and a 5.86 dB improvement over 3DGS~\cite{kerbl20233dgaussiansplattingrealtime}.
It also surpass baselines on SSIM and LPIPS matrices.   
This performance advantage stems from the fact that, as discussed earlier, the Jinc kernel is derived from the inverse Fourier transform of the 3D ideal low-pass filter, which drops to zero instantaneously at the cutoff frequency and theoretically suppresses aliasing completely. 
In contrast, the Gaussian kernel of 3DGS and the Student’s t-distribution kernel of SSS are both non-ideal low-pass filters, whose high-frequency leakage limits their performance. 
At $128\times128$ resolution, the Jinc kernel still maintains its lead in PSNR, which is 9.10 dB higher than that of Gaussians, continuing its advantage in the frequency domain. 
Overall, the Jinc kernel’s comprehensive dominance in low-resolution scenarios and its PSNR advantage in high-resolution scenarios confirm the effectiveness of designing an ideal low-pass reconstruction kernel from the perspective of signal processing to address aliasing issues in discrete sampling.

\begin{figure*}[t]
    \centering
    \includegraphics[width=\linewidth]{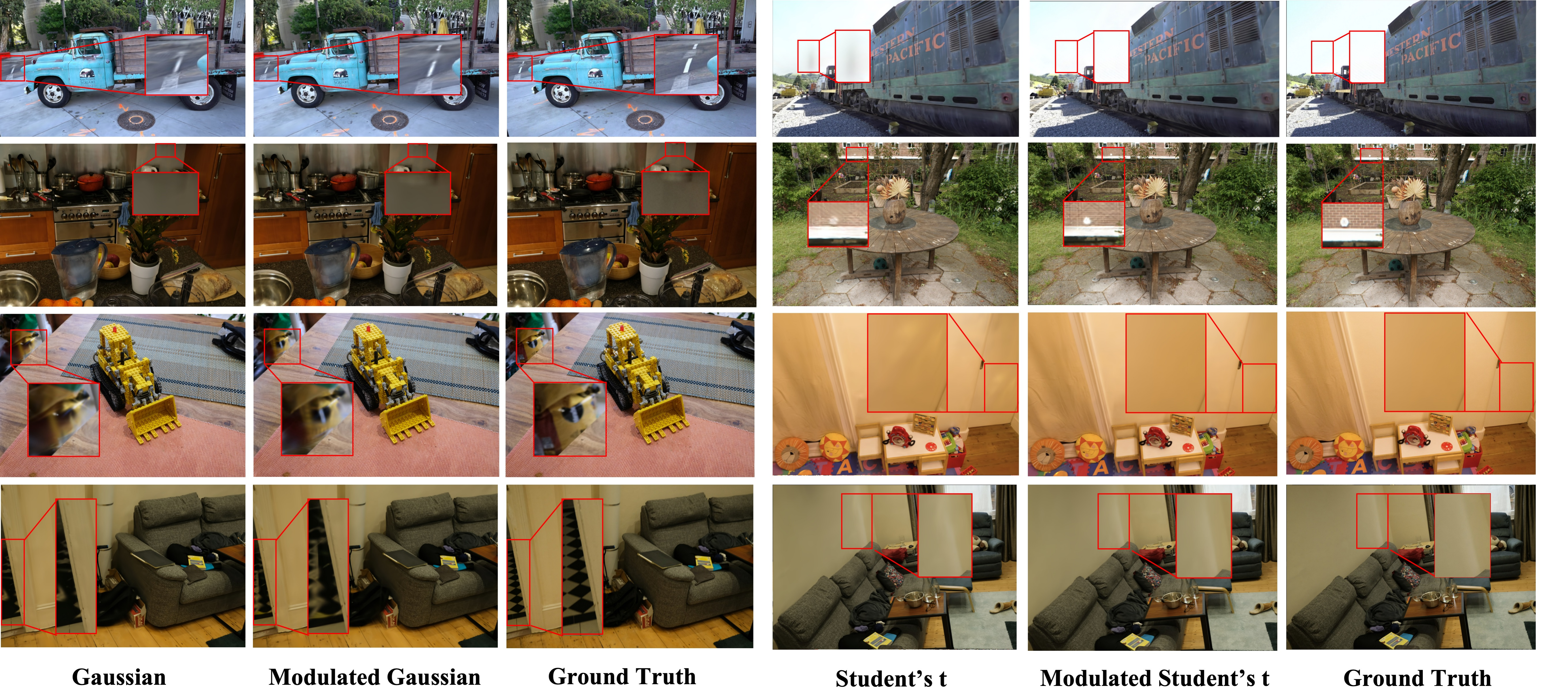}
    \caption{\textbf{Qualitative comparison.} We visualize the rendering results of based kernels~\cite{kerbl20233dgaussiansplattingrealtime, zhu20253dstudentsplattingscooping} and our modulated kernels.}
    \vspace{-0.2cm}
    \label{fig:visualization}
\end{figure*}

\subsection{Novel View Synthesis}

\textbf{Datasets.} To assess the effectiveness of our methods in novel view synthesis, we conduct experiments on both indoor and outdoor datasets, including 7 public scenes from Mip-NeRF 360~\cite{Barron_2022_CVPR}, two outdoor scenes from Tanks and Temples~\cite{TT2017ToG}, two indoor scenes from Deep Blending~\cite{DB2018ACM}.

\noindent\textbf{Baselines.} We compare our modulated methods with typical implicit neural rendering method Mip-NeRF360~\cite{Barron_2022_CVPR} and other Gaussian-based methods, including GES~\cite{GES2024CVPR}, 3DHGS~\cite{li20253dhgs3dhalfgaussiansplatting}, Fre-GS~\cite{FreGS2024CVPR}, Scaffold-GS~\cite{Scaffold-gs2024CVPR}, and 3DGS-MCMC~\cite{3DGS-MCMC}. Since our modulated methods are built on Gaussian and Student's t kernels, we also include 3DGS~\cite{kerbl20233dgaussiansplattingrealtime} and SSS~\cite{zhu20253dstudentsplattingscooping} for fair comparison.

\noindent\textbf{Implementation Details.} We implement our modulated reconstruction kernels on top of the original codebases of 3DGS~\cite{kerbl20233dgaussiansplattingrealtime} and SSS~\cite{zhu20253dstudentsplattingscooping}, respectively. For each baseline, we preserve the default training configurations provided in the corresponding codebase. Our modifications are limited strictly to the replacement of the original reconstruction kernel, ensuring that performance differences arise solely from the kernel design rather than changes in auxiliary components or hyperparameters.

\noindent\textbf{Main Results.} Tab.~\ref{tab:full resolution result} summarizes the quantitative evaluation on Mip-NeRF 360~\cite{Barron_2022_CVPR}, Tanks and Temples~\cite{TT2017ToG}, and Deep Blending~\cite{DB2018ACM}. Our modulation strategy yields consistent improvements for both 3DGS~\cite{kerbl20233dgaussiansplattingrealtime} and SSS~\cite{zhu20253dstudentsplattingscooping}. The performance gains on 3DGS~\cite{kerbl20233dgaussiansplattingrealtime} are particularly notable, reaching up to 0.72 dB PSNR and 0.012 SSIM, accompanied by lower LPIPS scores.
The SSS~\cite{zhu20253dstudentsplattingscooping} baseline also shows steady improvements across most settings, indicating that our modulation enhances both classical Gaussian kernels.
These trends can be explained by the frequency-domain characteristics shown in Fig.~\ref{fig:teaser}. For Gaussian kernels, modulation substantially strengthens the low-frequency response, which accounts for the large improvements observed on 3DGS~\cite{kerbl20233dgaussiansplattingrealtime}. However, for SSS~\cite{zhu20253dstudentsplattingscooping}, due to its slow decay in the frequency domain, the potential additional benefit gained from modulation is inherently smaller, leading to moderate yet still consistent performance gains. As shown in Fig.~\ref{fig:visualization}, the baseline methods~\cite{kerbl20233dgaussiansplattingrealtime,zhu20253dstudentsplattingscooping} struggle to capture background color consistency and fine structural details, often producing blurred textures. In contrast, our modulated kernels produce more accurate background colors and sharper scene details. These improvements align with the quantitative results in Table~\ref{tab:full resolution result}. Overall, the results confirm that modulation serves as an effective mechanism.

%% file: tab/low_resolution.tex
\begin{table}[t]
\centering
\caption{\textbf{Low resolution results.} We report PSNR, SSIM, and LPIPS of 3DGS~\cite{kerbl20233dgaussiansplattingrealtime}, SSS~\cite{zhu20253dstudentsplattingscooping}, and Jinc Splatting with $64\times64$ and $128\times128$ resolutions.}

\begin{tabular}{c|lccc}
\toprule
Resolution        & Methods & PSNR$\uparrow$  & SSIM$\uparrow$  & LPIPS$\downarrow$  \\ \midrule
\multirow{3}{*}{$64\times64$} & 3DGS~\cite{kerbl20233dgaussiansplattingrealtime}    & 24.01 & 0.901 & 0.0575 \\
                  & SSS~\cite{zhu20253dstudentsplattingscooping}     & 29.17 & 0.954 & 0.0214 \\
                  & Jinc    & 29.87 & 0.955 & 0.0199 \\ \midrule
\multirow{3}{*}{$128\times 128$} & 3DGS~\cite{kerbl20233dgaussiansplattingrealtime}    & 22.30  & 0.887  & 0.0956   \\
                  & SSS~\cite{zhu20253dstudentsplattingscooping}     & 30.24      & 0.964      & 0.0230       \\
                  & Jinc    & 31.40      & 0.961      & 0.0411       \\ \bottomrule
\end{tabular}

\label{tab:low resolution result}
\end{table}

%% file: tab/full_resolution.tex
\begin{table*}[t]
\centering
\caption{\textbf{Quantitative results of novel view synthesis.} We report matrices evaluated on Mip-NeRF360~\cite{Barron_2022_CVPR}, Tanks \& Temples~\cite{TT2017ToG}, and Deep Blending~\cite{DB2018ACM}.}
\setlength{\tabcolsep}{2.5mm}{\begin{tabular}{lccccccccc}
\toprule
\multirow{2}{*}{Method} & \multicolumn{3}{c}{Mip-NeRF360} & \multicolumn{3}{c}{Tanks \& Temples} & \multicolumn{3}{c}{Deep Blending} \\ \cmidrule{2-10}
                        & PSNR$\uparrow$     & SSIM $\uparrow$    & LPIPS $\downarrow$   & PSNR$\uparrow$       & SSIM$\uparrow$       & LPIPS$\downarrow$      & PSNR$\uparrow$      & SSIM$\uparrow$      & LPIPS$\downarrow$     \\ \midrule
Mip-NeRF~\cite{Barron_2021_ICCV}                & 29.23    & 0.844    & 0.207    & 22.22      & 0.759      & 0.257      & 29.40     & 0.901     & 0.245     \\
GES~\cite{GES2024CVPR}                     & 26.91    & 0.794    & 0.250    & 23.35      & 0.836      & 0.198      & 29.68     & 0.901     & 0.252     \\
3DHGS~\cite{li20253dhgs3dhalfgaussiansplatting}                   & 29.56    & 0.873    & 0.178    & 24.49      & 0.857      & 0.169      & 29.76     & 0.905     & 0.242     \\
Fre-GS~\cite{FreGS2024CVPR}                  & 27.85    & 0.826    & 0.209    & 23.96      & 0.841      & 0.183      & 29.93     & 0.904     & 0.240     \\
Scaffold-GS~\cite{Scaffold-gs2024CVPR}             & 28.84    & 0.848    & 0.220    & 23.96      & 0.853      & 0.177      & 30.21     & 0.906     & 0.254     \\
3DGS-MCMC~\cite{3DGS-MCMC}              & 29.89    & 0.900    & 0.190    & 24.29      & 0.860      & 0.190      & 29.67     & 0.890     & 0.320     \\ \midrule
\rowcolor{gray!8}
3DGS~\cite{kerbl20233dgaussiansplattingrealtime}                   & 28.69    & 0.870    & 0.182    & 23.14      &      0.841      & 0.183      & 29.41     & 0.903     & 0.243     \\
\rowcolor{gray!8}
\multirow[t]{2}{*}{Ours (3DGS)}  & \textbf{29.16}    & \textbf{0.871}    & \textbf{0.181}    & \textbf{23.86}      &      \textbf{0.853}      & \textbf{0.168}      & \textbf{29.84}     & \textbf{0.907}     & \textbf{0.237}        \\
\midrule
\rowcolor{gray!8}
SSS~\cite{zhu20253dstudentsplattingscooping}                   & 29.90    & \textbf{0.893}    & 0.145    & 24.87      &      0.873      & 0.138      & 30.07     & 0.907     & 0.247     \\
\rowcolor{gray!8}
\multirow[t]{2}{*}{Ours (SSS)}  & \textbf{29.96} & \textbf{0.893} & \textbf{0.143} & \textbf{24.95}  &  \textbf{0.874}   & \textbf{0.135} & \textbf{30.17}     & \textbf{0.911}     & \textbf{0.242}        \\
\bottomrule
\end{tabular}
}
\label{tab:full resolution result}
\end{table*}

%% file: sec/6_conclusion.tex
\section{Conclusion}
This paper revisits 3D reconstruction from a signal processing perspective, focusing on resolving aliasing caused by frequency-domain periodic extension in discrete sampling. We analyze existing reconstruction kernels (e.g., Gaussians, Student’s t-distributions) and identify their unideal low-pass limitations, which lead to high-frequency leakage.
To address this, we propose the Jinc kernel, which is derived from the 3D ideal low-pass filter’s inverse Fourier transform—that drops to zero at the cutoff frequency, theoretically eliminating aliasing. Based on this representations, we develop Jinc Splatting, including ray integral derivation, coordinate transformation, and gradient backpropagation for parameter optimization. 
To mitigate the Jinc kernel’s slow spatial decay, we introduce a frequency modulation strategy that combines its ideal frequency property with efficient spatial decay of existing kernels. 
Experimental results demonstrate the superiority of our Jinc and modulated kernels in novel view synthesis.

\noindent\textbf{Limitations and Future Works.}
Although we adopt a modulation strategy to alleviate the issues of the ideal Jinc kernel, the kernel itself still carries inherent limitations. Firstly, its slow spatial decay requires a large spatial support, leading to high memory consumption and visible rectangular artifacts. Secondly, the ideal kernel inherently exhibits ringing, producing oscillatory intensity transitions rather than a smooth monotonic falloff. These issues limit its practicality in scenarios that demand artifact-free representations. 
For the future works, the definition of new reconstruction kernels is still a challenging problem to achieve efficiency optimization as well as anti-aliasing.
Besides, the proposed modulation strategy could be extended to feed-forward reconstruction pipelines or 4D dynamic scenes, where improved anti-aliasing may further enhance temporal consistency and boost real-time rendering performance.
More attention can be focused on the generative models to fundamentally increase the sampling rates.

\noindent\textbf{Acknowledgments.} This work was supported in part by the National Natural Science Foundation of China under Grant 62576185, and by the Beijing Natural Science Foundation under Grant L252011.

%% file: sec/X_suppl.tex
\clearpage
\appendix
\maketitlesupplementary
\renewcommand\thesection{\Alph{section}}

\section{Jinc Kernel}
In this section, we perform detailed derivation of Jinc kernel proposed in Section~\ref{subsec: Jinc Kernel}.

\subsection{Frequency Domain Definition}
The ideal low-pass filter (ILPF) in three dimensions is a theoretical construct used in signal processing for multi-dimensional signals, such as volumetric data or 3D images. Its frequency domain transfer function is a spherical indicator function, allowing frequency components within a cutoff radius to pass unattenuated while blocking higher frequencies. The spatial domain response, or impulse response, is obtained via the inverse Fourier transform.
We use the following conventions:
\begin{align}
    H(\mathbf{f}) & = \iiint h(\mathbf{r}) e^{-j 2\pi \mathbf{f} \cdot \mathbf{r}} \, dx \, dy \, dz, \\
    h(\mathbf{r}) & = \iiint H(\mathbf{f}) e^{j 2\pi \mathbf{f} \cdot \mathbf{r}} \, df_x \, df_y \, df_z,
\end{align}
where $j = \sqrt{-1}$.

The transfer function of the 3D ideal low-pass filter with cutoff frequency \(f_c\) is:
\begin{equation}
    H(\mathbf{f}) = 
    \begin{cases} 
    1, & \|\mathbf{f}\| \leq f_c, \\
    0, & \|\mathbf{f}\| > f_c,
    \end{cases}
\end{equation}
where \(\|\mathbf{f}\| = \sqrt{f_x^2 + f_y^2 + f_z^2}\). This defines a spherical passband in the frequency domain.

\subsection{Derivation of the Spatial Response}
\begin{figure}
    \centering
    \includegraphics[width=\linewidth]{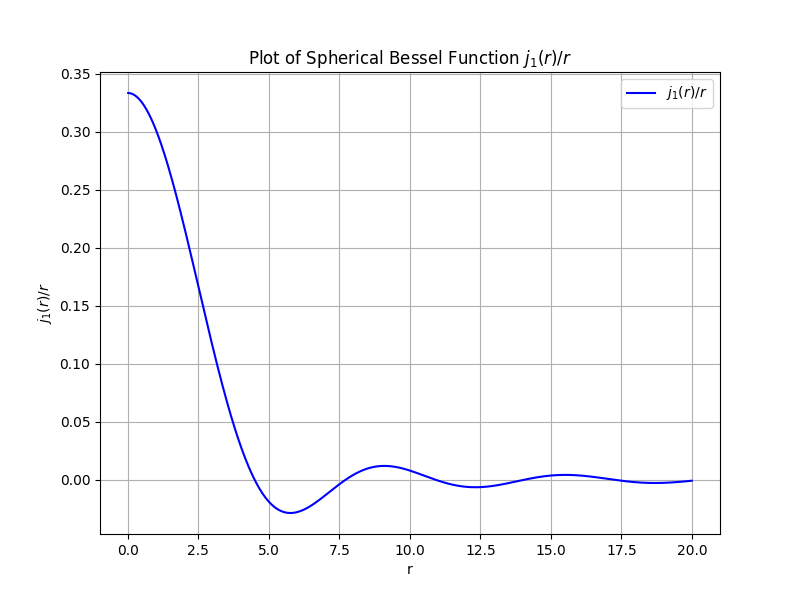}
    \caption{The sketch map of half of the ILPF.}
    \label{fig:ILPF}
\end{figure}

Due to the radial symmetry of \(H(\mathbf{f})\), the spatial response \(h(\mathbf{r})\) depends only on \(r = \|\mathbf{r}\| = \sqrt{x^2 + y^2 + z^2}\). We compute \(h(r)\) using spherical coordinates in the frequency domain.
The inverse Fourier transform becomes:
\begin{align}
    h(r) & = \int_0^{f_c} \int_0^{\pi} \int_0^{2\pi} e^{j 2\pi f r \cos\theta} f^2 \sin\theta \, d\phi \, d\theta \, df, \nonumber\\
    & = 2\pi \int_0^{f_c} f^2 \int_0^{\pi} e^{j 2\pi f r \cos\theta} \sin\theta \, d\theta \, df, \nonumber\\
    & = 2\pi \int_0^{f_c} f^2 \cdot \frac{2 \sin(2\pi f r)}{2\pi f r} \, df, \nonumber\\
    & = \frac{4\pi}{2\pi r} \int_0^{f_c} f^2 \cdot \frac{\sin(2\pi f r)}{f} \, df, \nonumber\\
    & = \frac{2}{r} \int_0^{f_c} f \sin(2\pi f r) \, df, \nonumber\\
    & = \frac{1}{2\pi^2 r^3} \left[ \sin(2\pi f_c r) - 2\pi f_c r \cos(2\pi f_c r)\right].
\end{align}
The spherical Bessel function of the first kind \(j_1(\alpha)\) is given as:
\begin{equation}
    j_1(\alpha) = \frac{\sin \alpha - \alpha \cos \alpha}{\alpha^2}.
\end{equation}
So, we have:
\begin{equation}
    h(r) = \frac{1}{2\pi^2 r^3} \cdot (2\pi f_c r)^2 j_1(2\pi f_c r) = \frac{2 f_c^2}{r} j_1(2\pi f_c r). \nonumber
\end{equation}
The curve of $j_1(r)/r$ is visualized in Figure~\ref{fig:ILPF}

\subsection{Anisotropy ILPF}
We have discussed about the isotropy ILPF above.
Now we introduce the definition of the anisotropy ILPF:
\begin{equation}
    h(\bm{x})=\dfrac{j_1(\sqrt{(\bm{x}-\bm{\mu})^{\top}\Sigma^{-1}(\bm{x}-\bm{\mu})})}{\sqrt{(\bm{x}-\bm{\mu})^{\top}\Sigma^{-1}(\bm{x}-\bm{\mu})}}, \label{eq:3D ILPF}
\end{equation}
where $\bm{\mu}$ is the center postion of ILPF and $\Sigma=RSS^{\top}R^{\top}$ is the covariance matrix.
Here we ignore the outoff frequency, which can be absorbed into the scale matrix $S$.

\section{Jinc Splatting}
In this section, we perform detailed derivation for Jinc Splatting in Section~\ref{subsec: Jinc Splatting}.

\subsection{Integration Along a Line}

This section derives the integral of the spatial response of a three-dimensional ideal low-pass filter along an arbitrary straight line \(\mathbf{x}(t) = \mathbf{a} + t \mathbf{b}\), where \(\mathbf{a} = (a_x, a_y, a_z)\), \(\mathbf{b} = (b_x, b_y, b_z)\), and \(t \in (-\infty, \infty)\).

We need to compute:
\begin{align}
    I & = \int_{-\infty}^{\infty} h(\bm{x}(t)) \, dt,  \nonumber\\
    & = \int_{-\infty}^{\infty} \dfrac{j_1(\sqrt{(\bm{x}(t)-\bm{\mu})^{\top}\Sigma^{-1}(\bm{x}(t)-\bm{\mu})})}{\sqrt{(\bm{x}(t)-\bm{\mu})^{\top}\Sigma^{-1}(\bm{x}(t)-\bm{\mu})}} \, dt, \nonumber\\
    & = \int_{-\infty}^{\infty} \dfrac{j_1\left(\sqrt{((\bm{a}-\bm{\mu})+t\bm{b})^{\top}\Sigma^{-1}((\bm{a}-\bm{\mu})+t\bm{b})}\right)}{\sqrt{((\bm{a}-\bm{\mu})+t\bm{b})^{\top}\Sigma^{-1}((\bm{a}-\bm{\mu})+t\bm{b})}} \, dt. \nonumber
\end{align}
Since $\Sigma=RSS^{\top}R^{\top}$, we define $\bm{m}=S^{-1}R^{-1}(\bm{a}-\bm{\mu})$ and $\bm{n}=S^{-1}R^{-1}\bm{b}$. Thus, we have:
\begin{align}
    &((\bm{a}-\bm{\mu})+t\bm{b})^{\top}\Sigma^{-1}((\bm{a}-\bm{\mu})+t\bm{b})\nonumber\\  = & (\bm{m}+t\bm{n})^{\top}(\bm{m}+t\bm{n}), \nonumber\\
    = &\Vert \bm{m}\Vert^{2}+2t(\bm{m}\cdot\bm{n}) + t^2\Vert \bm{n}\Vert^{2}.
\end{align}
Substitute $s = t+\bm{m}\cdot\bm{n}/\Vert \bm{n}\Vert^{2}$, the integration transfers to:
\begin{align}
    I & = \int_{-\infty}^{\infty} \dfrac{j_1\left(\sqrt{((\bm{a}-\bm{\mu})+t\bm{b})^{\top}\Sigma^{-1}((\bm{a}-\bm{\mu})+t\bm{b})}\right)}{\sqrt{((\bm{a}-\bm{\mu})+t\bm{b})^{\top}\Sigma^{-1}((\bm{a}-\bm{\mu})+t\bm{b})}} \, dt, \nonumber \\
    & = \int_{-\infty}^{\infty} \dfrac{j_1\left(\sqrt{\Vert \bm{m}\Vert^{2}+2t(\bm{m}\cdot\bm{n}) + t^2\Vert \bm{n}\Vert^{2}}\right)}{\sqrt{\Vert \bm{m}\Vert^{2}+2t(\bm{m}\cdot\bm{n}) + t^2\Vert \bm{n}\Vert^{2}}} \, dt, \nonumber \\
    & = \int_{-\infty}^{\infty} \dfrac{j_1\left(\Vert \bm{n}\Vert\sqrt{s^{2}+\dfrac{\Vert\bm{m}\times\bm{n}\Vert^2}{\Vert\bm{n}\Vert^4}}\right)}{\Vert \bm{n}\Vert\sqrt{s^{2}+\dfrac{\Vert\bm{m}\times\bm{n}\Vert^2}{\Vert\bm{n}\Vert^4}}} \, ds, \nonumber \\
    & = \int_{-\infty}^{\infty} \dfrac{j_1\left(\dfrac{\Vert\bm{m}\times\bm{n}\Vert}{\Vert\bm{n}\Vert}\sqrt{v^{2}+1}\right)}{\Vert\bm{n}\Vert\sqrt{v^{2}+1}}\, dv, \nonumber\\
    & = \int_{-\infty}^{\infty} \dfrac{j_1\left(\alpha\sqrt{v^2+1}\right)}{\Vert\bm{n}\Vert\sqrt{v^2+1}}\, dv, \nonumber \\
    & = \dfrac{\pi J_1(\alpha)}{\Vert\bm{n}\Vert\alpha}
\end{align}

\subsection{Gradient Backward Propagation}
This section derives the backward propagation for Jinc Splatting in Section~\ref{subsubsec: Gradient}.
\subsubsection{The Partial Derivative of $I$ with respect to $\alpha$}
First, we have the differential of $j_1(x)$:
\begin{equation}
    j_1'(x)=j_0(x)-\frac{2}{x}j_1(x),
\end{equation}
\begin{equation}
    j_0(x)=\frac{sin(x)}{x},\quad j_1(x)=\frac{sin(x)-xcos(x)}{x^2}.
\end{equation}
Thus:
\begin{align}
    \frac{\partial I}{\partial\alpha}
    & = \int_{-\infty}^{\infty} \frac{1}{\Vert \bm{n}\Vert\sqrt{v^2+1}}\frac{\partial j_1\left(\alpha\sqrt{v^2+1}\right)}{\partial\alpha}\, dv \nonumber\\
    & = \int_{-\infty}^{\infty} \frac{\partial j_1\left(\alpha\sqrt{v^2+1}\right)}{\Vert \bm{n}\Vert\partial\left(\alpha\sqrt{v^2+1}\right)}\, dv \nonumber\\
    & = \frac{1}{\Vert \bm{n}\Vert}\int_{-\infty}^{\infty} j_0\left(\alpha\sqrt{v^2+1}\right)-\frac{2j_1\left(\alpha\sqrt{v^2+1}\right)}{\alpha\sqrt{v^2+1}}\, dv \nonumber\\
    & = \frac{1}{\Vert \bm{n}\Vert}\left[\dfrac{\pi J_0(\alpha)}{\alpha}-\dfrac{2\pi J_1(\alpha)}{\alpha^2}\right].
\end{align}
\subsubsection{The Partial Derivative of $\alpha$ with respect to $\mu$}
The partial derivative of $\alpha$ with respect to $\mu$ can be formulated as:
\begin{align}
    \frac{\partial \alpha}{\partial \bm{\mu}}
    &= \frac{1}{2\alpha}
       \frac{\partial}{\partial\bm{\mu}}
       \left(
           \|\bm{m}\|^2
           - \frac{(\bm{m}^\top \bm{n})^2}{\|\bm{n}\|^2}
       \right) \nonumber\\
    &= \frac{1}{2\alpha}
       \left(
            \frac{\partial (\bm{m}^\top \bm{m})}{\partial\bm{\mu}}
            - \frac{\partial}{\partial\bm{\mu}}
              \left(
                  \frac{(\bm{m}^\top \bm{n})^2}{\|\bm{n}\|^2}
              \right)
       \right) \nonumber\\
    &= \frac{1}{2\alpha}
       \left(
            2\left(\frac{\partial \bm{m}}{\partial\bm{\mu}}\right)^\top \bm{m}
            - \frac{2(\bm{m}^\top \bm{n})}{\|\bm{n}\|^2}
              \left(\frac{\partial \bm{m}}{\partial\bm{\mu}}\right)^\top \bm{n}
       \right) \nonumber\\
    &= \frac{1}{2\alpha}
       \left(
            -2(\bm{S}^{-1}\bm{R}^{-1})^\top \bm{m}
            + \frac{2(\bm{m}^\top \bm{n})}{\|\bm{n}\|^2} (\bm{S}^{-1}\bm{R}^{-1})^\top \bm{n}
       \right) \nonumber\\
    &= \frac{1}{\alpha}
       \left(
            -(\bm{S}^{-1}\bm{R}^{-1})^\top \bm{m}
            + \frac{\bm{m}^\top \bm{n}}{\|\bm{n}\|^2} (\bm{S}^{-1}\bm{R}^{-1})^\top \bm{n}
       \right) \nonumber\\
    &= -\frac{1}{\alpha}(\bm{S}^{-1}\bm{R}^{-1})^\top
       \left(
           \bm{m} - \frac{\bm{m}^\top \bm{n}}{\|\bm{n}\|^2} \bm{n}
       \right) \nonumber\\
    &= -\frac{1}{\alpha}(\bm{RS})^{-\top}
       \left(
           \bm{m} - \frac{\bm{m}^\top \bm{n}}{\|\bm{n}\|^2} \bm{n}
       \right).
\end{align}
\subsubsection{The Partial Derivative of $\alpha$ with respect to $\Sigma$}
The partial derivative of $\alpha$ with respect to $\Sigma$ can be formulated as:
\begin{align}
    \frac{\partial\alpha}{\partial\bm{\Sigma}}
    & = \frac{1}{2\alpha}(\frac{\partial (\bm{m}^\top \bm{m})}{\partial\bm{\Sigma}}-\frac{\partial}{\partial\bm{\Sigma}}\left(
                  \frac{(\bm{m}^\top \bm{n})^2}{\|\bm{n}\|^2}
              \right)).
\end{align}
We introduce the standard form of matrix differential in the numerator layout convention:
\begin{equation}
df=\text{tr}((\frac{\partial f}{\partial\mathbf{X}})^\top\mathbf{X})
\end{equation}
Similar to the partial derivative of $\alpha$ with respect to $\mu$, given the standard form of matrix differential, we can get that:
\begin{align}
    & \frac{\partial \alpha}{\partial \bm{\Sigma}} = \frac{\bm{\Sigma}^{-1}}{2\alpha (\bm{n}^\top \bm{\Sigma}^{-1} \bm{n})^2} \left( - (\bm{n}^\top \bm{\Sigma}^{-1} \bm{n})^2 \bm{m} \bm{m}^\top \right.\\
& \left. + 2 (\bm{m}^\top \bm{\Sigma}^{-1} \bm{n})(\bm{n}^\top \bm{\Sigma}^{-1} \bm{n}) \bm{n} \bm{m}^\top - (\bm{m}^\top \bm{\Sigma}^{-1} \bm{n})^2 \bm{n} \bm{n}^\top \right) \bm{\Sigma}^{-1}. \nonumber
\end{align}
\input{tab/nerf360_psnr}
\input{tab/nerf360_ssim}
\input{tab/nerf360_lpips}
\section{Additional Experimental Results}
\subsection{Detailed Results on Each Scene}
\input{tab/tant_db_psnr}
\input{tab/tant_db_ssim}
\input{tab/tant_db_lpips}
We report detailed results on every scene in Mip-NeRF 360~\cite{Barron_2022_CVPR}, Tanks $\text{\&}$ Temples~\cite{TT2017ToG} and Deep Blending~\cite{DB2018ACM} for our modulated methods. We choose Mip-NeRF~\cite{Barron_2021_ICCV}, 3DGS~\cite{kerbl20233dgaussiansplattingrealtime}, Scaffold-GS~\cite{Scaffold-gs2024CVPR}, 3DHGS~\cite{li20253dhgs3dhalfgaussiansplatting}, 3DGS-MCMC~\cite{3DGS-MCMC} and SSS~\cite{zhu20253dstudentsplattingscooping} as baselines.

As shown in Tabs.~\ref{tab: nerf360_psnr} to~\ref{tab: tant_db_lpips}, in general, our modulation strategy outperforms both 3DGS~\cite{kerbl20233dgaussiansplattingrealtime} and SSS~\cite{zhu20253dstudentsplattingscooping}. While the modulation strategy may not produce positive improvements in every scene, it delivers clear and consistent benefits in the majority cases.
\subsection{Multi-Resolution Analysis}
To further validate the correctness of our theoretical insights, we conduct a comprehensive multi-resolution evaluation. As the ideal jinc kernel exhibits slow spatial decay, a practical implementation inevitably requires spatial truncation. We therefore compare the truncated jinc kernel with the standard Gaussian kernel used in 3DGS by training both methods at the native resolution, followed by evaluation at the original resolution, as well as several downsampled resolutions. The experiment is conducted in Tanks \& Temples~\cite{TT2017ToG}.
\input{tab/multi_resolution}

The results shown in Tab.~\ref{tab:multi-resolution result} demonstrate a consistent and theoretically aligned trend. At the native resolution, the truncated jinc kernel exhibits a slight performance drop relative to the Gaussian baseline, which is expected given the truncation. However, as the test resolution decreases, the spatial support required for accurate reconstruction becomes correspondingly smaller. Under these conditions, the jinc kernel more faithfully preserves the frequency characteristics of the underlying signal and outperforms the Gaussian kernel. This resolution-dependent behavior provides additional evidence for the correctness of our theoretical analysis.

%% file: tab/nerf360_psnr.tex
\begin{table*}[t]
\centering
\renewcommand{\arraystretch}{0.85}
\caption{\textbf{PSNR results for every scene in Mip-NeRF 360~\cite{Barron_2022_CVPR}
dataset}.}
\setlength{\tabcolsep}{2.5mm}{\begin{tabular}{l|ccccccc|c}
\toprule
Method & bicycle & bonsai & counter & garden & kitchen & room & stump & average \\\midrule
Mip-NeRF~\cite{Barron_2021_ICCV} & 24.37 & 33.46 & 29.55 & 26.98 & 32.23 & 31.64 & 26.40 & 29.23  \\
3DHGS~\cite{li20253dhgs3dhalfgaussiansplatting} & 25.39 & 33.30 & 29.62 & 27.68 & 32.17 & 32.12 & 26.64 & 29.56\\
Scaffold-GS~\cite{Scaffold-gs2024CVPR} & 24.50 & 32.70 & 29.34 & 27.17 & 31.30 & 31.93 & 26.27 & 28.84\\
3DGS-MCMC~\cite{3DGS-MCMC} & 26.15 & 32.88 & 29.51 & 28.16 & 32.27 & 32.48 & 27.80 & 29.89\\ \midrule
\rowcolor{gray!8}
3DGS~\cite{kerbl20233dgaussiansplattingrealtime} & \textbf{25.25} & 31.98 & 28.70 & \textbf{27.41} & 30.32 & 30.63 & 26.55 & 28.69\\
\rowcolor{gray!8}
\multirow[t]{2}{*}{Ours (3DGS)} & 25.17 & \textbf{32.18} & \textbf{29.19} & \textbf{27.41} & \textbf{31.69} & \textbf{31.78} & \textbf{26.69} & \textbf{29.16}   \\
\midrule
\rowcolor{gray!8}
SSS~\cite{zhu20253dstudentsplattingscooping} & 25.68 & 33.50 & 29.87 & 28.09 & 32.43 & 32.57 & 27.17 & 29.90\\
\rowcolor{gray!8}
\multirow[t]{2}{*}{Ours (SSS)}  & \textbf{25.76} & \textbf{33.59} & \textbf{29.89} & \textbf{28.12} & \textbf{32.52} & \textbf{32.65} & \textbf{27.21} & \textbf{29.96}   \\
\bottomrule
\end{tabular}
}
\label{tab: nerf360_psnr}
\end{table*}

%% file: tab/nerf360_ssim.tex
\begin{table*}[t]
\centering
\renewcommand{\arraystretch}{0.85}
\caption{\textbf{SSIM results for every scene in Mip-NeRF 360~\cite{Barron_2022_CVPR}
dataset}.}
\setlength{\tabcolsep}{2.5mm}{\begin{tabular}{l|ccccccc|c}
\toprule
Method & bicycle & bonsai & counter & garden & kitchen & room & stump & average \\\midrule
Mip-NeRF~\cite{Barron_2021_ICCV} & 0.685 & 0.941 & 0.894 & 0.813 & 0.920 & 0.913 & 0.744 & 0.844  \\
3DHGS~\cite{li20253dhgs3dhalfgaussiansplatting} & 0.768 & 0.950 & 0.909 & 0.868 & 0.930 & 0.921 & 0.770 & 0.873 \\
Scaffold-GS~\cite{Scaffold-gs2024CVPR} & 0.705 & 0.946 & 0.914 & 0.842 & 0.928 & 0.925 & 0.784 & 0.848\\
3DGS-MCMC~\cite{3DGS-MCMC} & 0.810 & 0.950 & 0.920 & 0.890 & 0.940 & 0.940 & 0.820 & 0.900\\ \midrule
\rowcolor{gray!8}
3DGS~\cite{kerbl20233dgaussiansplattingrealtime} & \textbf{0.771} & 0.938 & 0.905 & \textbf{0.868} & 0.922 & 0.914 & \textbf{0.775} & 0.870 \\
\rowcolor{gray!8}
\multirow[t]{2}{*}{Ours (3DGS)} & 0.749 & \textbf{0.946} & \textbf{0.915} & 0.858 & \textbf{0.933} & \textbf{0.928} & 0.769 & \textbf{0.871}   \\
\midrule
\rowcolor{gray!8}
SSS~\cite{zhu20253dstudentsplattingscooping} & \textbf{0.798} & \textbf{0.956} & \textbf{0.926} & 0.882 & 0.939 & 0.938 & \textbf{0.813} & \textbf{0.893}\\
\rowcolor{gray!8}
\multirow[t]{2}{*}{Ours (SSS)}  & \textbf{0.798} & 0.954 & 0.925 & \textbf{0.883} & \textbf{0.940} & \textbf{0.939} & \textbf{0.813} & \textbf{0.893}   \\
\bottomrule
\end{tabular}
}
\label{tab: nerf360_ssim}
\end{table*}

%% file: tab/nerf360_lpips.tex
\begin{table*}[t!]
\centering
\renewcommand{\arraystretch}{0.85}
\caption{\textbf{LPIPS results for every scene in Mip-NeRF 360~\cite{Barron_2022_CVPR}
dataset}.}
\setlength{\tabcolsep}{2.5mm}{\begin{tabular}{l|ccccccc|c}
\toprule
Method & bicycle & bonsai & counter & garden & kitchen & room & stump & average \\\midrule
Mip-NeRF~\cite{Barron_2021_ICCV} & 0.301 & 0.176 & 0.204 & 0.170 & 0.127 & 0.211 & 0.261 & 0.207  \\
3DHGS~\cite{li20253dhgs3dhalfgaussiansplatting} & 0.202 & 0.180 & 0.201 & 0.104 & 0.125 & 0.220 & 0.215 & 0.178 \\
Scaffold-GS~\cite{Scaffold-gs2024CVPR} & 0.306 & 0.185 & 0.191 & 0.146 & 0.126 & 0.202 & 0.284 & 0.220  \\
3DGS-MCMC~\cite{3DGS-MCMC} & 0.180 & 0.220 & 0.220 & 0.100 & 0.140 & 0.250 & 0.190 & 0.190 \\ \midrule
\rowcolor{gray!8}
3DGS~\cite{kerbl20233dgaussiansplattingrealtime} & \textbf{0.205} & 0.205 & 0.204 & \textbf{0.103} & 0.129 & 0.220 & \textbf{0.210} & 0.182 \\
\rowcolor{gray!8}
\multirow[t]{2}{*}{Ours (3DGS)} & 0.239 & \textbf{0.177} & \textbf{0.181} & 0.121 & \textbf{0.115} & \textbf{0.193} & 0.242 & \textbf{0.181}   \\
\midrule
\rowcolor{gray!8}
SSS~\cite{zhu20253dstudentsplattingscooping} & \textbf{0.173} & 0.151 & 0.156 & \textbf{0.090} & 0.104 & 0.167 & \textbf{0.174} & 0.145\\
\rowcolor{gray!8}
\multirow[t]{2}{*}{Ours (SSS)}  & \textbf{0.173} & \textbf{0.141} & \textbf{0.155} & 0.091 & \textbf{0.103} & \textbf{0.166} & \textbf{0.174} & \textbf{0.143}   \\
\bottomrule
\end{tabular}
}
\label{tab: nerf360_lpips}
\end{table*}

%% file: tab/tant_db_psnr.tex
\begin{table*}[t]
\centering
\renewcommand{\arraystretch}{0.85}
\caption{\textbf{PSNR results for every scene in Tanks \& Temples~\cite{TT2017ToG} and Deep Blending~\cite{DB2018ACM}.}}
\setlength{\tabcolsep}{2.5mm}{\begin{tabular}{l|ccc|ccc}
\toprule
\multirow{2}{*}{Method} & \multicolumn{3}{c}{Tanks \& Temples} & \multicolumn{3}{c}{Deep Blending} \\ \cmidrule{2-7}
                        & train     & truck    & average   & drjohnson       & playroom       & average     \\ \midrule
Mip-NeRF~\cite{Barron_2021_ICCV} & 19.52 & 24.91 & 22.22 & 29.14 & 29.66 & 29.40       \\
3DHGS~\cite{li20253dhgs3dhalfgaussiansplatting} & 22.95 & 26.04 & 24.49 & 29.32 & 30.20 & 29.76        \\
Scaffold-GS~\cite{Scaffold-gs2024CVPR} & 22.15 & 25.77 & 23.96 & 29.80 & 30.62 & 30.21  \\
3DGS-MCMC~\cite{3DGS-MCMC} & 22.47 & 26.11 & 24.29 & 29.00 & 30.33 & 29.67     \\ \midrule
\rowcolor{gray!8}
3DGS~\cite{kerbl20233dgaussiansplattingrealtime} & 21.09 & 25.18 & 23.14 & 28.77 & 30.04 & 29.41     \\
\rowcolor{gray!8}
\multirow[t]{2}{*}{Ours (3DGS)} & \textbf{22.18} & \textbf{25.54} & \textbf{23.86} & \textbf{29.41} & \textbf{30.26} & \textbf{29.84}  \\
\midrule
\rowcolor{gray!8}
SSS~\cite{zhu20253dstudentsplattingscooping} & 23.32 & 26.41 & 24.87 & 29.66 & 30.47 & 30.07  \\
\rowcolor{gray!8}
\multirow[t]{2}{*}{Ours (SSS)} & \textbf{23.47} & \textbf{26.42} & \textbf{24.95} & \textbf{29.71} & \textbf{30.62} & \textbf{30.17}      \\
\bottomrule
\end{tabular}
}
\label{tab: tant_db_psnr}
\end{table*}

%% file: tab/tant_db_ssim.tex
\begin{table*}[t]
\centering
\renewcommand{\arraystretch}{0.85}
\caption{\textbf{SSIM results for every scene in Tanks \& Temples~\cite{TT2017ToG} and Deep Blending~\cite{DB2018ACM}.}}
\setlength{\tabcolsep}{2.5mm}{\begin{tabular}{l|ccc|ccc}
\toprule
\multirow{2}{*}{Method} & \multicolumn{3}{c}{Tanks \& Temples} & \multicolumn{3}{c}{Deep Blending} \\ \cmidrule{2-7}
                        & train     & truck    & average   & drjohnson       & playroom       & average     \\ \midrule
Mip-NeRF~\cite{Barron_2021_ICCV} & 0.660 & 0.857 & 0.759 & 0.901 & 0.900 & 0.901       \\
3DHGS~\cite{li20253dhgs3dhalfgaussiansplatting} & 0.827 & 0.887 & 0.857 & 0.904 & 0.907 & 0.905        \\
Scaffold-GS~\cite{Scaffold-gs2024CVPR} & 0.822 & 0.883 & 0.853 & 0.907 & 0.904 & 0.906  \\
3DGS-MCMC~\cite{3DGS-MCMC} & 0.830 & 0.890 & 0.860 & 0.890 & 0.900 & 0.890
    \\ \midrule
\rowcolor{gray!8}
3DGS~\cite{kerbl20233dgaussiansplattingrealtime} & 0.802 & 0.879 & 0.841 & 0.899 & 0.906 & 0.903     \\
\rowcolor{gray!8}
\multirow[t]{2}{*}{Ours (3DGS)} & \textbf{0.820} & \textbf{0.885} & \textbf{0.853} & \textbf{0.905} & \textbf{0.908} & \textbf{0.907}  \\
\midrule
\rowcolor{gray!8}
SSS~\cite{zhu20253dstudentsplattingscooping} & 0.850 & \textbf{0.897} & 0.873 & 0.905 & 0.909 & 0.907  \\
\rowcolor{gray!8}
\multirow[t]{2}{*}{Ours (SSS)} & \textbf{0.851} & \textbf{0.897} & \textbf{0.874} & \textbf{0.908} & \textbf{0.914} & \textbf{0.911}     \\
\bottomrule
\end{tabular}
}
\label{tab: tant_db_ssim}
\end{table*}

%% file: tab/tant_db_lpips.tex
\begin{table*}[t!]
\centering
\renewcommand{\arraystretch}{0.85}
\caption{\textbf{LPIPS results for every scene in Tanks \& Temples~\cite{TT2017ToG} and Deep Blending~\cite{DB2018ACM}.}}
\setlength{\tabcolsep}{2.5mm}{\begin{tabular}{l|ccc|ccc}
\toprule
\multirow{2}{*}{Method} & \multicolumn{3}{c}{Tanks \& Temples} & \multicolumn{3}{c}{Deep Blending} \\ \cmidrule{2-7}
                        & train     & truck    & average   & drjohnson       & playroom       & average     \\ \midrule
Mip-NeRF~\cite{Barron_2021_ICCV} & 0.354 & 0.159 & 0.257 & 0.237 & 0.252 & 0.245       \\
3DHGS~\cite{li20253dhgs3dhalfgaussiansplatting} & 0.197 & 0.141 & 0.169 & 0.240 & 0.243 & 0.242        \\
Scaffold-GS~\cite{Scaffold-gs2024CVPR} & 0.206 & 0.147 & 0.177 & 0.250 & 0.258 & 0.254  \\
3DGS-MCMC~\cite{3DGS-MCMC} & 0.240 & 0.140 & 0.190 & 0.330 & 0.310 & 0.320\\ \midrule
\rowcolor{gray!8}
3DGS~\cite{kerbl20233dgaussiansplattingrealtime} & 0.218 & 0.148 & 0.183 & 0.244 & 0.241 & 0.243    \\
\rowcolor{gray!8}
\multirow[t]{2}{*}{Ours (3DGS)} & \textbf{0.196} & \textbf{0.140} & \textbf{0.168} & \textbf{0.235} & \textbf{0.238} & \textbf{0.237}   \\
\midrule
\rowcolor{gray!8}
SSS~\cite{zhu20253dstudentsplattingscooping} & 0.166 & 0.109 & 0.138 & 0.249 & 0.245 & 0.247  \\
\rowcolor{gray!8}
\multirow[t]{2}{*}{Ours (SSS)} & \textbf{0.163} & \textbf{0.107} & \textbf{0.135} & \textbf{0.246} & \textbf{0.239} & \textbf{0.242}      \\
\bottomrule
\end{tabular}
}
\label{tab: tant_db_lpips}
\end{table*}

%% file: tab/multi_resolution.tex
\begin{table*}[t!]
\centering
\caption{\textbf{Quantitative results with different resolutions in Tanks \& Temples~\cite{TT2017ToG}.} }
\setlength{\tabcolsep}{2.0mm}{\begin{tabular}{l|ccc|ccc|ccc|ccc}
\toprule
\multirow{2}{*}{Method} & \multicolumn{3}{c}{Full Resolution} & \multicolumn{3}{c}{1/2 Resolution} & \multicolumn{3}{c}{1/4 Resolution} & \multicolumn{3}{c}{1/8 Resolution} \\ \cmidrule{2-13}
& PSNR     & SSIM    & LPIPS   & PSNR       & SSIM       & LPIPS      & PSNR      & SSIM      & LPIPS      &
PSNR       & SSIM       & LPIPS       \\ \midrule
3DGS~\cite{kerbl20233dgaussiansplattingrealtime} & \textbf{23.14} & \textbf{0.841} & \textbf{0.183} & \textbf{22.16} & \textbf{0.838} & \textbf{0.136} & 18.39 & \textbf{0.686} & 0.195 & 15.77 & 0.538 & 0.252 \\
Jinc & 22.55 & 0.824 & 0.186 & 21.98 & 0.812 & 0.143 & \textbf{18.91} & 0.673 & \textbf{0.169} & \textbf{16.47} & \textbf{0.559} & \textbf{0.204} \\
\bottomrule
\end{tabular}
}
\label{tab:multi-resolution result}
\end{table*}